\documentclass[10pt,twocolumn,letterpaper]{article}

\usepackage[pagenumbers]{cvpr} %

\usepackage{makecell} 
\usepackage[table]{xcolor}
\newcommand{\greyrule}{\arrayrulecolor{black!30}\midrule\arrayrulecolor{black}}

\definecolor{gray}{HTML}{aaaaaa}
\definecolor{lightblue}{HTML}{acd0f6}
\definecolor{blue}{HTML}{5c94b5}
\definecolor{green}{HTML}{b6e288}
\definecolor{yellow}{HTML}{f0f567}
\definecolor{red}{HTML}{f85f5f}

\usepackage{comment}

\definecolor{cvprblue}{rgb}{0.21,0.49,0.74}
\usepackage[pagebackref,breaklinks,colorlinks,citecolor=cvprblue]{hyperref}
\usepackage{float}

\def\paperID{12996} %
\def\confName{CVPR}
\def\confYear{2025}

\title{Neural Image Unfolding: \\Flattening Sparse Anatomical Structures using Neural Fields}

\author{Leonhard Rist$^{1,2}$, ~ Pluvio Stephan$^{1}$, ~ Noah Maul$^{1}$, ~ Linda Vorberg,$^{1,2}$\\Hendrik Ditt$^{2}$, ~ Michael Sühling$^{2}$, ~ Andreas Maier$^{1}$, ~ Bernhard Egger$^{3}$, ~ Oliver Taubmann$^{2}$\\
\normalsize $^1$Pattern Recognition Lab, Friedrich-Alexander-Universität Erlangen-Nürnberg, Germany\\ \normalsize $^2$Computed Tomography, Siemens~Healthineers~AG, Forchheim, Germany\\\normalsize $^3$Chair of Visual Computing, Friedrich-Alexander-Universität Erlangen-Nürnberg, Germany\\
{\tt\small leonhard.rist@fau.de}
}

\begin{document}

\twocolumn[{%
\renewcommand\twocolumn[1][]{#1}%
\maketitle
\begin{center}
    \centering
    \captionsetup{type=figure}
    \includegraphics[width=\textwidth]{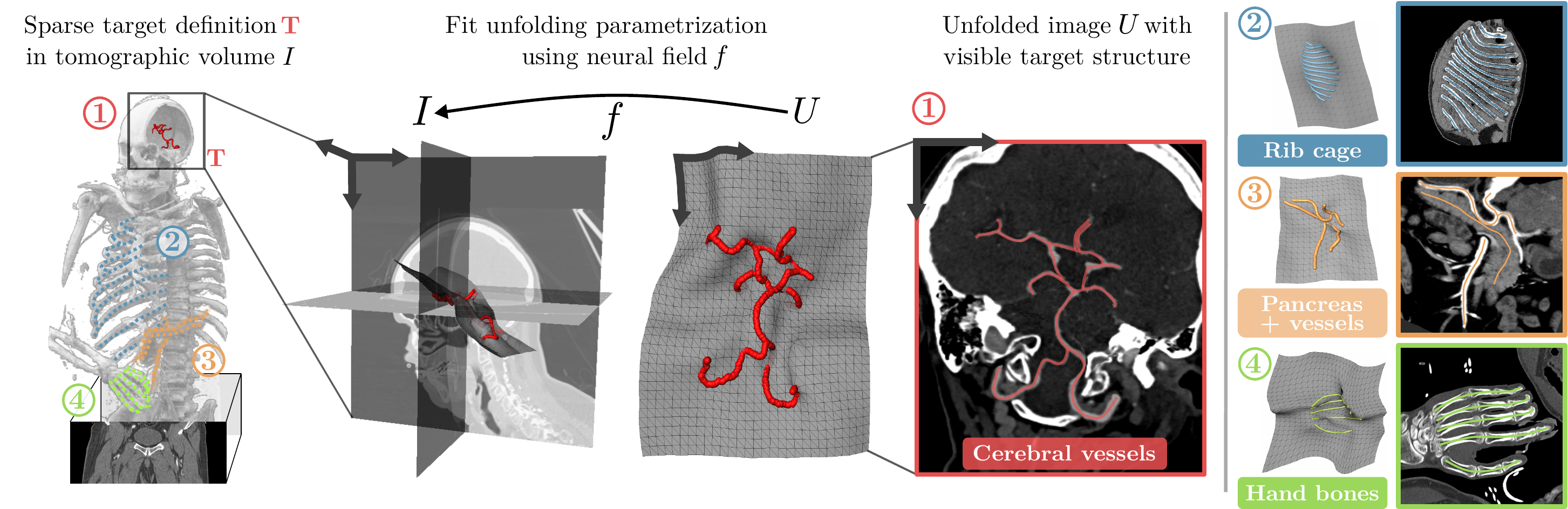}
    \captionof{figure}{Unfolding sparse cerebral vessels using a neural field $f$: Fitting a 2-dimensional manifold in the tomographic image $I$ from a sparse set of automatically extracted 3D target centerline points $\mathbf{T}$ (red). Three unfolding examples on the right with depicted centerlines.} 
    \label{fig:teaser}
\end{center}%
}]

\begin{abstract}

Tomographic imaging reveals internal structures of 3D objects and is crucial for medical diagnoses.  
Visualizing the morphology and appearance   
of non-planar sparse anatomical structures that extend over multiple 2D slices in tomographic volumes is inherently difficult but valuable for decision-making and reporting.
Hence, various organ-specific unfolding techniques exist to map their densely sampled 3D surfaces to a distortion-minimized 2D representation.
However, there is no versatile framework to flatten complex sparse structures including vascular, duct or bone systems.
We deploy a neural field to fit the transformation of the anatomy of interest to a 2D overview image. %
We further propose distortion regularization strategies and
combine geometric with intensity-based loss formulations to also display non-annotated and auxiliary targets.
In addition to improved versatility, our unfolding technique outperforms mesh-based baselines for sparse structures w.r.t.~%
peak distortion and our regularization scheme yields smoother transformations compared to Jacobian formulations from neural field-based image registration.
\end{abstract}

\section{Introduction}
\label{sec:intro}

Acquiring volumetric images using non-destructive tomographic scanning enables the investigation of internal structures, which is heavily leveraged for medical diagnostics using X-ray Computed Tomography (CT) or Magnetic Resonance Imaging.  
The resulting 3D volumes are virtually sliced to obtain multiple 2D views in specific orientations, which may only visualize a fraction of the object of interest (e.g.~a blood vessel cross section).
Alternatively employed 3D volume rendering techniques~\cite{Zhang2010}  
have similar limitations due to viewing angle and occlusions, rendering standardized and canonical visual representations difficult.

Flattening or unfolding a 2D surface manifold comprising all target points (e.g.~vessel centerline positions) into a planar representation as shown in Fig.~\ref{fig:teaser} helps to simultaneously inspect the morphology of a topological structure (e.g.~bifurcations) and its appearance (e.g.~to assess calcification or stenosis).
Such standardized views are helpful for disease tracking, inter-patient comparison, volume navigation and reporting or as lower-dimensional representations for other downstream tasks. %
Hence, many applications of medical unfolding exist (Sec.~\ref{sec:medical_unfolding}) with the goal of finding a distortion-minimized parametrization of a targeted organ surface, reminiscent of the common computer vision problem of mapping textures to 3D objects (Sec.~\ref{sec:classicalMeshPara}).
However, there is a lack of flexible unfolding techniques for thin and sparse structures, which also require the definition of such a surface in addition to its optimal parametrization.
Many of the required objectives could be formulated as optimization problems and neural fields -- small multi-layer perceptrons to model coordinate transforms -- have emerged as an efficient way to fit practically useful solutions (Sec.~\ref{sec:nf_literature}). 

We propose a simple but powerful, modality-independent neural field approach to unfold sparse 3D objects into 2D representations without requiring global point ordering. 
For this purpose, we fit 2D manifolds to such sparse point sets in a distortion-minimized manner and employ multiple extensions for more effective and efficient fitting.
The contribution of this manuscript is threefold:
\begin{enumerate}
    \item We present a versatile and customizable framework comprised of a pointwise-fitted neural field to unfold a 2D manifold containing a sparse target point set. 
    \item We introduce a novel multi-scale distance distortion regularizer, which goes hand in hand with the resolution independence of the neural field.
    \item We combine geometric with image-based loss formulations to jointly optimize for desirable appearance criteria and/or auxiliary targets in the image domain.%
\end{enumerate}

\noindent Exemplarily, we present our results for four relevant tasks: Rib unfolding to detect small tumors or fractures, hand unfolding that can visualize degenerative disease for patients unable to open their hand, 
unfolding the pancreas and its surrounding arteries for cancer resection planning and cerebral vessel unfolding to support clinicians in stroke assessment.

\begin{figure*}[tb]
\centering
\includegraphics[width=\linewidth]{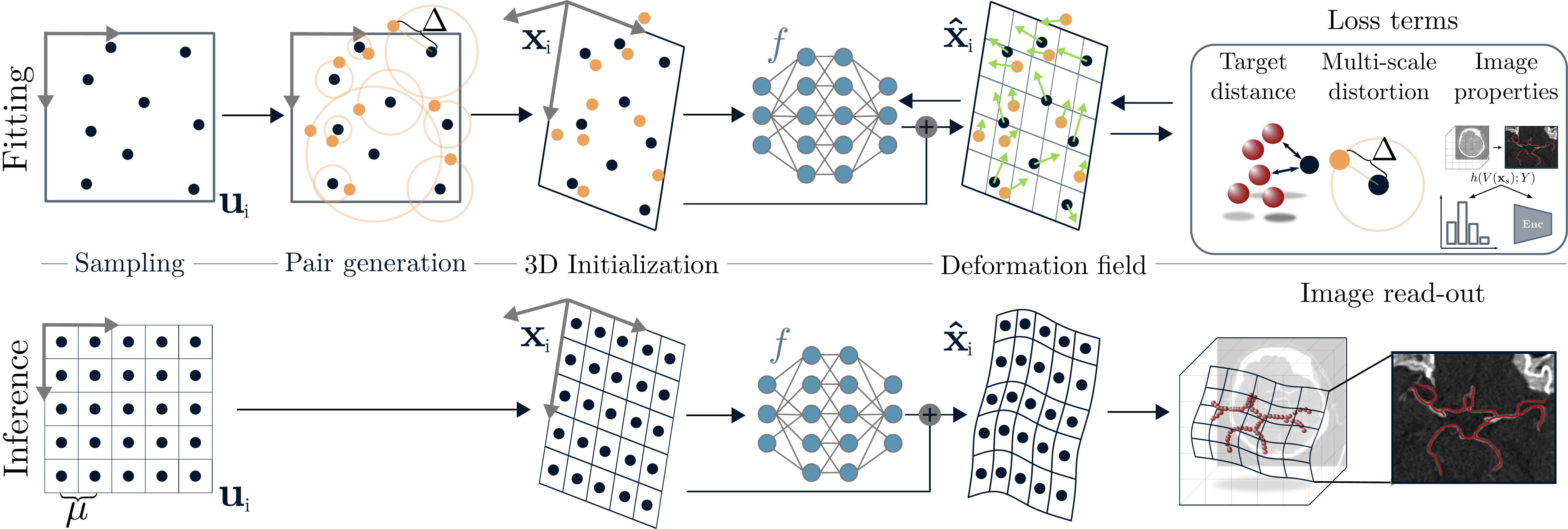}
\caption{\textbf{Neural unfolding pipeline} for the fitting process using randomly sampled points (top) and for inference using a grid input leading to a structured image after read-out (bottom).}
\label{fig:pipeline}
\end{figure*}

\section{Related Work}

\subsection{Classical Mesh Parameterization}
\label{sec:classicalMeshPara}
3D mesh parameterization for texture mapping or remeshing is a fundamental computer vision problem~\cite{MeshParamOverview_Sheffer} extending to point cloud parameterization~\cite{PointCloudParametrization}. 
Its goal is a (bijective) coordinate mapping of an object that generates a 2D representation with minimal distortion, either conformal~\cite{confmap_texture,mullen2008spectral}, authalic~\cite{desbrun2002intrinsic} or -- as in our medical scenario -- isometric. 
Classical texture mapping approaches range from free boundaries and distortion minimizing energies~\cite{hormann2000mips} over patchwise progressive meshes~\cite{progressiveMesh} including fold-preventing barrier functions~\cite{smith2015bijective} to intrinsic parametrizations~\cite{ desbrun2002intrinsic}.
Shape-preserving parametrization such as the local/global approach~\cite{local_global} maps mesh triangles individually and integrates known mesh deformation methods such as the as-rigid-as possible (ARAP) algorithm~\cite{Arap} to distribute the vertices.
ARAP iteratively minimizes the local rigidity energy using local rotations to fit a mesh to a target point set.
However, these algorithms require densely sampled surfaces, are difficult to adapt to specific use cases and cannot readily include intensity-based constraints.

\subsection{Unfolding Techniques in Medical Imaging}

\label{sec:medical_unfolding}

The multitude of different unfolding techniques for medical applications~\cite{survey_medicalUnfolding} highlights their immense value for diagnostics, ranging from schematic plots~\cite{termeer2007covicad, BEP} over 2D surface~\cite{cortex1,colon} and centerline mappings~\cite{cpr} to volumetric unfolding~\cite{placenta}. %
They are often designed for a specific organ such as the brain cortex~\cite{cortex1} or colon~\cite{colon} and require their densely sampled surface. 
Other methods exploit similarity to geometric primitives such as the sphericity of the skull~\cite{curvedMIP_skull} or by using raycasting, e.g.~for the rib cage~\cite{ribunfolding} to find tumors or tiny fractures. %
Anatomy-driven reformation (ADR)~\cite{adr} generalizes unfolding for curved surfaces, parameterizing initialized meshes with the local/global~\cite{local_global} ARAP optimization. 
Abulnaga et al.~tackle the 3D unfolding of the placenta with Dirichlet energies~\cite{placenta}.
For sparse structures such as the vasculature, curved planar reformation (CPR) is the most widely used unfolding technique~\cite{cpr}. %
CPR traverses sorted centerline points to fill each image row and %
includes straightening, multi-path, untangled and spiral extensions~\cite{cpr,spiralcpr}, which are, however, unable to unfold disjoint or circular structures and can exhibit artifacts at strongly curved segments.
These shortcomings are addressed by CeVasMap~\cite{cevasmap} for the intracranial arteries by fitting splines along the principal axes of a sparse point set to initialize a read-out mesh, followed by ARAP and image assembly for ill-conditioned vessel combinations.
In conclusion, sparse targets (centerlines or individual landmarks) are of high clinical relevance, yet most prior work requires densely sampled surfaces or is limited to tree-based structures. 
While CeVasMap can handle arbitrary point sets, it focuses only on the target geometry, can exhibit local distortions and introduces discontinuities due to image assembly.

\subsection{Neural Fields in Geometric Processing}
\label{sec:nf_literature}

Neural feature representations for geometry processing include Neural Surface maps~\cite{NeuralSurfaceMap} or parametric encodings~\cite{MeshFeat}, can be processed by TextureNet~\cite{TextureNet} or are used to extract surfaces from implicit representations~\cite{nerfmeshing,DeepMesh}. 
Flattening-Net~\cite{FlattenNet} parameterizes point clouds into a planar 2D representation but is incompatible with human perception.
DA Wand~\cite{liu2023wand} extracts local sub regions for parametrization whereas the Flatten-Anything-Model~\cite{FAM} uses multiple sub-networks to model the whole mesh parameterization process for densely sampled surfaces.
Implicit neural representations have also found medical applications~\cite{molaei2023implicit}, e.g.~to perform super-resolution~\cite{MR_superres}, segmentation~\cite{NF_segmentation} or clean reconstruction in 3D microscopy~\cite{kniesel2022clean}.
Neural fields for image registration model properties similar to our surface-to-target-point deformation.
Wolterink et al.~\cite{wolterink2022implicit} deform an image pointwise using a SIREN~\cite{SIREN} network with Jacobian and higher-order derivative regularizers to prevent folding.
Similarly, Sun et al.~\cite{NIR} fit their registration displacement field using global and patch-wise sampling schemes while a follow-up work uses velocity fields for the diffeomorphic registrations~\cite{DiffeomorphicIR_NVF}, which can also be solved using a Neural ODE solver (NODEO~\cite{wu2022nodeo}).

\section{Objective}
\label{sec:objectives}

We target a set of $N$ points $\mathbf{T}$\,=\,$\{\mathbf{t}_1, \mathbf{t}_2,..., \mathbf{t}_N \mid \mathbf{t}_n$$\in$$\mathbb{R}^{3} \}$, which should be visible in an unfolded image $U$.
Each intensity value $U({\mathbf{u}_i)}$ for a 2D position $\mathbf{u}_i \in [0,1]^2$ from a continuous, normalized image coordinate system corresponds to an intensity value $I(\mathbf{\hat{x}}_i)$ at $\mathbf{\hat{x}}_i \in \mathbb{R}^3$ in the tomographic volume with $U(\mathbf{u}_i)$\,=\,$ I(\mathbf{\hat{x}}_i)$.
$U$ 
is formed by evaluating a set of (regularly sampled) image positions during inference. %
For this purpose, we first transform the sampled points $\mathbf{u}_i$ into 3D with an initial heuristic 2D-3D transformation $\mathbf{x}_i$\,=\,$i(\mathbf{u}_i)$, 
such that the subsequent transformation $f$ only needs to model their displacement, similarly as  in image registration~\cite{wolterink2022implicit}.
Consequently, the volumetric read-out position $\mathbf{\hat{x}}_i$ is obtained by $ \mathbf{\hat{x}}_i$\,=\,$f(\mathbf{x}_i) + \mathbf{x}_i $, see Fig.~\ref{fig:pipeline}.
We optimize $f$ using geometric and intensity-based objectives:    %
\begin{enumerate}
    \item $L_{\text{t}}$: Outputs $\mathbf{\hat{x}}_i$ shall be as close as possible to the targets $\mathbf{T}$ such that marked structures are visible in the image. 
    \item $L_{\text{d}}$: Since pixels are equidistant, corresponding 3D distances shall also be constant to avoid image distortion.
    \item $L_{\text{im}}$: To maximize relevant image content beyond primary targets, we optionally introduce secondary objectives defined by volumetric or intensity-based metrics.
\end{enumerate}

\noindent These objectives are weighted and combined as
\begin{equation}
    L = w_{\text{t}}L_{\text{t}} + w_{\text{d}}L_{\text{d}} + w_{\text{im}} L_{\text{im}}\,.
\end{equation}

\section{Unfolding Quality Measures}
\label{sec:losses}
Below, the objectives from Sec.~\ref{sec:objectives} are specified as loss function terms with suitable optimization strategies for fitting $f$. 

\subsection{Distance to Target}
The target points $\mathbf{t}_n$ should have minimal distance to the implicit surface defined by $\mathbf{\hat{x}}_i$. 
Common template strategies use predefined target to pixel correspondences that, however, discard clinically relevant curvature information. %
Hence, a more flexible problem formulation focusing on the minimal distance of target to output coordinates 
could enable smoother deformations that result in more natural-looking unfoldings through seamless integration with other loss functions.
Inspired by Chamfer distance formulations, our target loss over a batch of predictions $S$ is the average distance of $\mathbf{T}$ to the closest transformed point (approximating the distance to the implicit surface) and reads
\begin{equation}
    L_{\text{t}} = \frac{1}{N} \sum^N_{n=1} \min_{s \in S}||\mathbf{\hat{x}}_s - \mathbf{t}_n||^2\,.
\end{equation}
Despite the non-differentiability of the $\min$ operation, fitting behaved robustly in all our experiments compared to smoother variants, such as top-k-$\min$ or $\operatorname{softmin}$.

\subsection{Image Distortion}
\label{sec:dist_loss}
Let $\Delta$ denote the distance between any two pixels $\mathbf{u}_i$ and $\mathbf{u}_j$. 
A deviation from the distance $\Delta$ between their corresponding 3D points leads to distortion
\begin{equation}
    d(\mathbf{\hat{x}}_i,\mathbf{\hat{x}}_j) = ||\mathbf{\hat{x}}_i -\mathbf{\hat{x}}_j|| - \Delta,
\end{equation}
where  $d$\,$>$\,0 indicates contraction and $d$\,$<$\,0 stretching. 
Hence, preserving this neighborhood isometry during projection of non-planar 3D subspaces to 2D mitigates distortion, although global isometry is unattainable in most cases.
For image quality measurements, only direct neighbors are typically considered by setting $\Delta = \delta$ to compute the overall distortion with $\delta$ being the distance of direct neighbors in case of isotropic pixel spacing.

To optimize transformations independent of an underlying grid, we propose the loss definition  
\begin{equation}
    L_{\text{d}} = \frac{1}{S} \sum_{s=1}^S w_s \cdot \left(||\mathbf{\hat{x}}_{s,1} - \mathbf{\hat{x}}_{s,2}|| - ||\mathbf{u}_{s,1} - \mathbf{u}_{s,2}||\right)^2,
    \label{eq:distortion}
\end{equation}
which is evaluated for jointly generated pairs of points $(\mathbf{u}_{s,1}, \mathbf{u}_{s,2})$ and corresponding predictions $(\mathbf{\hat{x}}_{s,1}, \mathbf{\hat{x}}_{s,2})$, with $w_s$ being an optional weighting factor discussed in Sec.~\ref{sec:importancemap}.
During fitting, we generate a second point per input with a random angle to and random distance $\Delta$ from the original point. 
This leads to a sampling at different scales, as seen in Fig.~\ref{fig:pipeline}, favoring isometry for the given range of distances and thus smoothing the resulting manifold.
The same approach could also be readily applied to further geometric properties of choice other than isometric distortion, e.g.~conformality. %

\subsection{Image-based Enhancement}
\label{sec:im_based_loss}
As we target structures that only occupy a sparse subspace in the unfolded image, the remaining (target-free) image area is, apart from the smoothness constraint, unrestricted and likely of little diagnostic value.
Thus, it is valuable to optimize for any desirable appearance property that guides the use of these target-free areas with image-based loss terms. %
For instance in vessel unfolding, where annotations can be incomplete or focused on main vessels, we can encourage the unfolding of smaller non-annotated branches or retrieve segments with missing annotations with the same intensity profile.
Similarly, the occupied image area of a secondary volumetric target (e.g.~described by an organ mask) can be maximized. 
For loss calculation, a read-out of the transformed coordinates $\mathbf{\hat{x}}_i$ in a volume $V$ is compared to an expected value $Y$ using a weighting function $h$ such that
\begin{equation}
    L_\text{im} = \sum_{s=1}^S h(V(\mathbf{\hat{x}}_s), Y). %
\end{equation}
$V$ can be equivalent to $I$ for intensity comparisons or a mask image, while $h$ is task specific, 
e.g.~a reward for vessel intensities or a classification loss for masks.
In principle, incorporation of image appearance can now be readily extended to the (additional) optimization of other downstream tasks to favor specific appearance properties that may be described analytically or by proximity of embeddings of a suitable image encoder (cf.~perceptual losses).

\section{Neural Fields for Unfolding}
\label{sec:nfu}

Neural fields are chosen as function approximators for the fitted 3D deformation $f$ due to their efficiency in representing transformations and flexibility in resolution-independent modeling. 
To obtain the final unfolded image during inference, uniformly distributed coordinate positions (pixels) are initially mapped to 3D space (Sec.~\ref{sec:initialization}), then deformed by $f$ to intersect the target, followed by a read-out of the volumetric intensity space $I$, see Fig.~\ref{fig:pipeline}. 
The sampling and weighting of pixel positions (Sec.~\ref{sec:importancemap}) and their embedding (Sec.~\ref{sec:input}) 
is described in the following.
In our experiments, the neural field consists of three hidden layers with width 128 and Leaky ReLU activations. %

\subsection{Initialization}
\label{sec:initialization}
To generate an initial target-centered 3D subspace, we elevate the sampled 2D input space linearly into 3D by arranging the first two vectors of the principal component analysis (PCA) of the target points in a matrix $\mathbf{A}$ to transform the normalized points $\mathbf{u}_i$ into PCA-space $\mathbf{x}_i = \mathbf{H} \cdot \mathbf{A}\mathbf{u}_i+\mathbf{t}_{\text{mean}}$.
To ensure proper coverage, we scale this linear subspace with a margin in addition to the target extent using the diagonal 3D matrix $\mathbf{H}$. 
This subspace, following the initial plane formulation from CeVasMap \cite{cevasmap}, is then deformed by $f$.
\begin{figure}[tp]
    \centering
    \includegraphics[width=\linewidth]{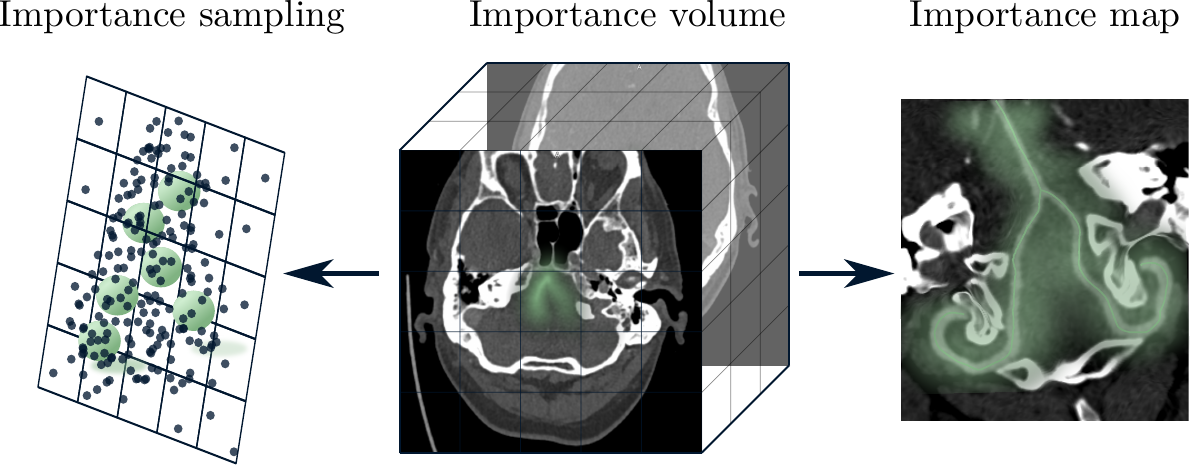}
    \caption{\textbf{Importance sampling} (left) \textbf{and map} overlay (right) from an importance definition in volumetric space which is inversely proportional to the Euclidean distance to the target (green).}
    \label{fig:importance_map}
\end{figure}

\begin{figure*}[tb]
    \centering
    \includegraphics[width=\linewidth]{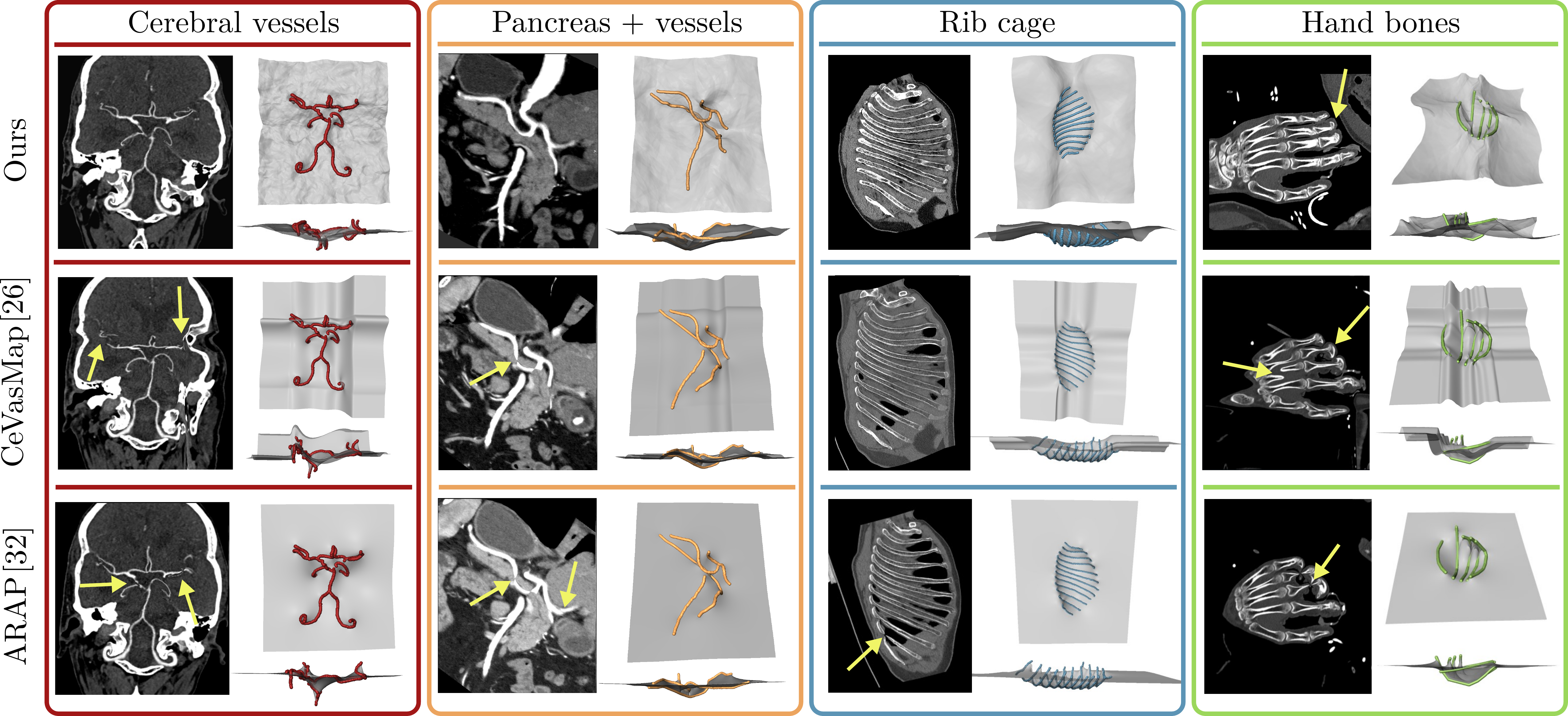}
    \caption{\textbf{Neural unfolding results} (first row) \textbf{compared to two baselines}, CeVasMap with spline predeformation (center) \cite{cevasmap} and with planar initialization equivalent to an ARAP deformation (bottom row) \cite{Arap}. Unfolded images are presented together with their read-out mesh showing two different views. The target points shown with the mesh are depicted as image overlays in Fig.~\ref{fig:teaser} for improved clarity. Important aspects and artifacts are highlighted in yellow. All images have an isotropic image resolution of 0.5\,mm.}
    \label{fig:res:baseline}
\end{figure*}

\subsection{Importance Sampling \& Map}
\label{sec:importancemap}
Input sampling strategies for fitting neural fields in image registration include uniform \cite{wolterink2022implicit} or downscaled (global) plus patchwise (local) schemes \cite{NIR}. 
To accelerate fitting at relevant and more complex locations, we assign less importance to transformations distant to the targets.
Proximity to target positions can be precalculated by applying the Euclidean Distance Transform $e(I)$ to the volume grid. 
A volume of importance weights $V_\text{E} = \frac{|\min(e(I) - \alpha, 0)| + \beta}{\alpha + \beta}$ inverts the distance map and clips values $>$\,$\alpha$, %
giving more weight to points close to the target while assigning a minimum value of $\beta$ to the rest.
Transforming a coarse input grid with the neural field and reading out $V_\text{E}$ yields a 2D importance map $E$, as shown in Fig.~\ref{fig:importance_map}.
We perform importance sampling by selecting points according to the distribution in $E$, while ensuring coverage of all regions with $\beta$\,$>$\,$0$. %
Inspired by ADR~\cite{adr}, one can utilize the value of $V_{\text{E}}(\mathbf{\hat{x}}_i)$ during model fitting as weighting factor $w_s$ in Eq.~\eqref{eq:distortion}.
This emphasizes small distortions close to the target and 
can help to move inevitable distortions to less important areas. 
We choose a fixed sampling size of 50k point pairs per epoch that are optimized in a single batch.

\subsection{Embedding \& Normalization}
\label{sec:input}

Following common practice, we embed our input using trigonometric functions.
In contrast to the various frequency components in images, neighboring positions of smooth displacement vector fields are highly correlated.
This suggests a low-frequency bias for the unfolding deformation.
Therefore, we choose a low number of 3 frequencies for the embedding which proved sufficient.
The network's 3D input is represented in the PCA-vector-based coordinate system of the target.
To improve generalization w.r.t.~scaling effects for the trigonometric functions, we perform a fixed normalization by dividing $\mathbf{\hat{x}}_i$ by a constant $c$ dependent on the target extent. %

\section{Experiments}
\label{sec:experiments}

For our experiments, we perform four relevant unfolding tasks of various sizes and complexity and compare our results to mesh-based baselines that can inherently handle sparse targets, namely CeVasMap \cite{cevasmap} and ARAP \cite{Arap} using the same initial plane description. 
We evaluate 10 patients for cerebral vessel unfolding (used in the ablation studies) and 5 cases for the other tasks, respectively. 
Data specifics are reported in Sup.~Tab.~\ref{tab:sup:dataset}. 
In an ablation study, we present visual examples with image distortion and target-to-mesh distance metrics for the respective important pipeline steps.
We compare our multi-scale distortion regularizer to Jacobian-based methods %
and investigate the image-based loss for object retrieval as well as to unfold an auxiliary volumetric target.  %
Finally, we measure the influence of importance weighting and sampling for image quality and optimization convergence. 
Distortion values based on Eq.~\eqref{eq:distortion} are reported in the relevant area around the unfolded target (1\,cm radius) to be independent of the mesh size.
All images have an isotropic image resolution of 0.5\,mm.

\subsection{Applications in CT Imaging}

\begin{figure}
    \centering
    \includegraphics[width=.95\linewidth]{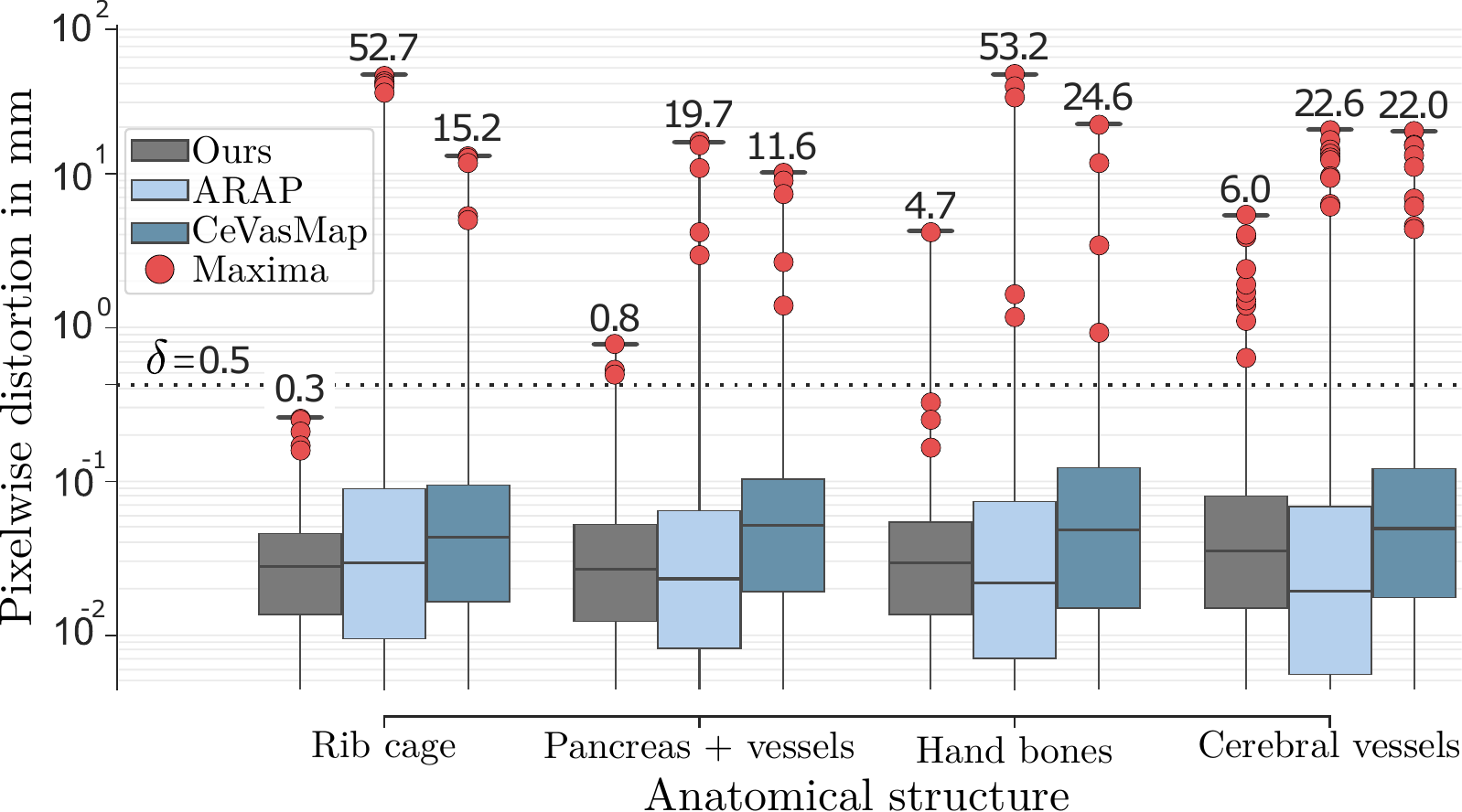}
    \caption{\textbf{Logarithmic distortion distribution} as boxplot (median line, std-dev.~boxes, maximum whiskers) within 1\,cm radius around unfolded target for each anatomy and method. Horizontal line of pixel spacing $\delta$, approx.~indicating the beginning of severe distortions. Red dots represent max.~distortion per case.}
    \label{fig:res:dist_plot}
\end{figure}

Unfolded images for all four applications are shown in Fig.~\ref{fig:res:baseline}. 
The resulting surfaces do not exhibit folding, are smooth and follow the object centerlines, see metrics in Sup.~Tab.~\ref{tab:sup:baseline}.
All models are fitted using the same configuration (multi-scale regularizer, weak importance map, importance sampling, loss weights $w_{\text{t}}$\,=\,2, $w_{\text{d}}$=1, $w_{\text{im}}$=0) and are optimized using stochastic gradient descent for 5000 epochs, requiring 86\,ms/step (Nvidia RTX A2000) leading to several minutes of fitting.  
ARAP deformation takes between 8\,s (410$\times$336 mesh) and 105\,s (1297$\times$1014 mesh).
The meshes for all tasks are deformed in target-distant areas to account for central deformations, most prominently seen for the ribs. 
The scenario of the bent hand requires coverage of the palm and fingertips, which is a similar geometric constellation as for the pancreas and vessel centerline. 
This is resolved by forming global sickle-like deformations to cover parallel structures, however missing the last part of the tips.
The high complexity in the center of the multiple cerebral vessel tree branches is also depicted morphologically correct.
We compare ourselves to CeVasMap \cite{cevasmap} with its proposed spline initialization as well as with a plane (same as ours), reducing it to a ARAP deformation \cite{Arap}. %
ARAP and CeVasMap struggle with structures that exhibit high curvature and where two non-adjacent points have the same closest point on the plane. 
This causes those points to coincide in the image, as seen in the cerebral vessels, finger joints and pancreas (yellow arrows in Fig.~\ref{fig:res:baseline}). 
They tend to displace mesh vertices mostly along their normal direction resulting in high local distortions for distant targets. %
This is reflected in the relevant max.~distortion values in Fig.~\ref{fig:res:dist_plot} which are consistently higher for CeVasMap and ARAP, corresponding to the highlighted artifacts in Fig.~\ref{fig:res:baseline}.
CeVasMap can avoid some of these high distortions with its prefitting.
Yet, the parallel structures of the hand can lead to counter-intuitive pre-deformation that results in visually disrupted structures. 
All methods exhibit inevitable but sufficiently low mean distortions smaller than 0.1\,mm. 
When investigating target distances, see Sup.~Tab.~\ref{tab:sup:baseline}, CeVasMap and ARAP show smaller mean values than our neural field by enforcing the target intersection. 
However, we are still below $\delta$ (except for large rib cage) and hence sample the targeted structures. 

\subsection{Distortion Regularization}

\begin{figure}[tb]
    \centering
    \includegraphics[width=0.94\linewidth]{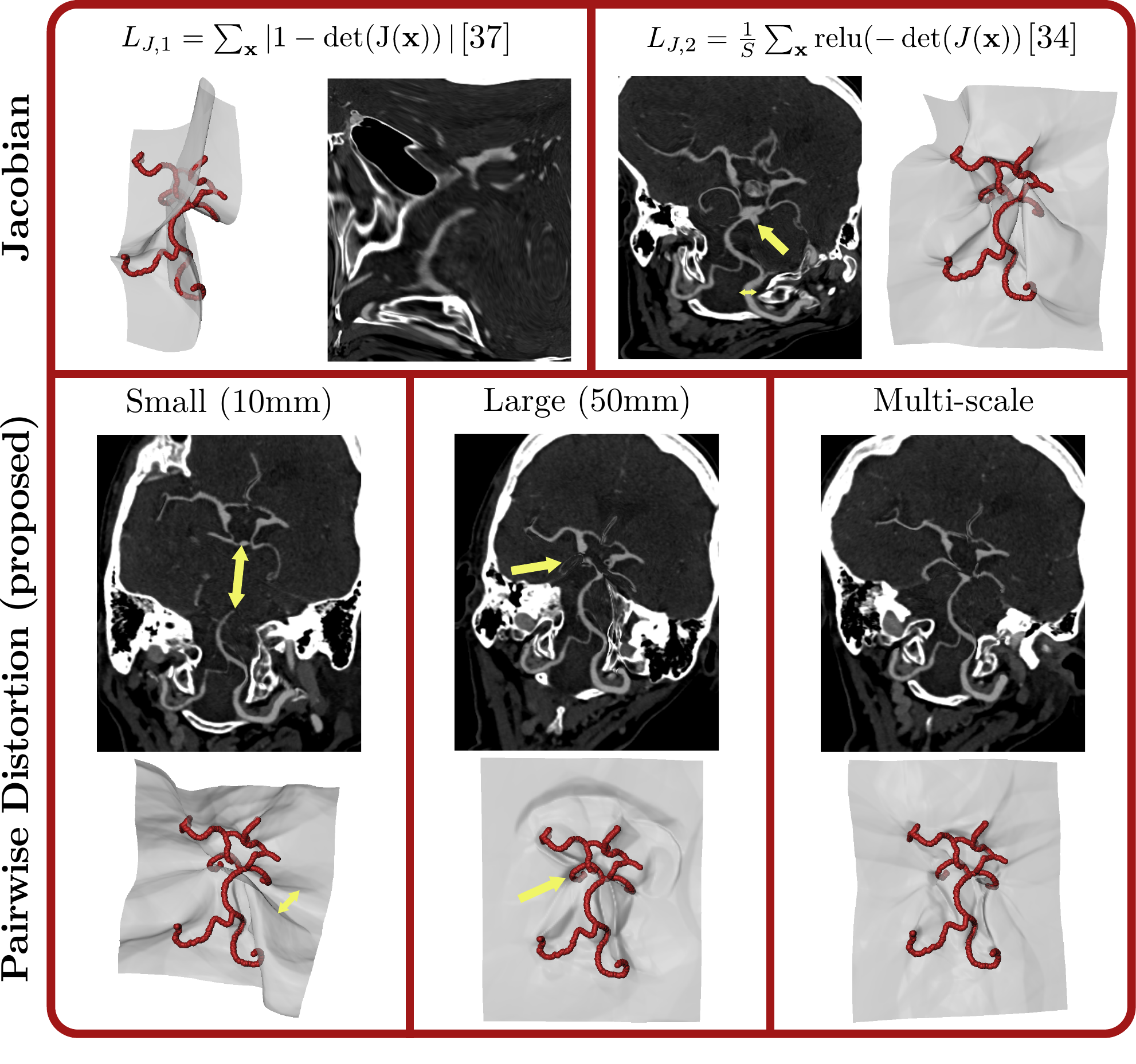}
    \caption{\textbf{Distortion regularizers} image and mesh results for cerebral vessels. Two Jacobian regularizers (top) \cite{wolterink2022implicit, DiffeomorphicIR_NVF} vs.~our neighborhood regularization (bottom) with small, large and multi-scale (0.5-40\,mm) distances. Artifacts are highlighted in yellow.}
    \label{fig:res:distortion}
\end{figure}

We choose the multi-branch cerebrovascular task as the most complex example to quantitatively evaluate ablations on 10 cases.
The influence of distortion regularizers is illustrated in Fig.~\ref{fig:res:distortion} and evaluated as above with distortion statistics reported in Sup.~Tab.~\ref{tab:sup:dist_ablations}.
The Jacobian regularizer $L_{J,1} = \sum_{\mathbf{x}} |1-\operatorname{det(J(\mathbf{x}))}|$ \cite{wolterink2022implicit} prevents overlapping and introduces little local distortion (max.~0.96\,mm), but heavily changes the initial plane structure leading to an unsuitable representation. %
Wolterink et al.~additionally employ a hyperelastic regularizer and a bending energy term that is infeasible in our case due to LeakyReLU activations. 
The regularizer $L_{J,2} = \frac{1}{S} \sum_{\mathbf{x}} \operatorname{relu}(-\operatorname{det}(J(\mathbf{x}))$ by Sun et al.~\cite{NIR} is within 1\,mm distance for over 97\,\% of the target points in Fig.~\ref{fig:res:distortion}, leading to a reasonable morphology out of the box. 
However, some vessel radii are stretched, resulting in blurry borders, %
whereas the same appear sharper and isometric with our proposed regularization. %

With a fixed small distance between point pairs, isometry between pixel neighbors is preserved. %
Nevertheless, (non-overlapping) bulges can form, stretching the image. 
This introduces undesired geodesic distances causing an incorrect representation of the target morphology, even though only little neighborhood distortion occurs. %
In our example, the fold in the center of Fig.~\ref{fig:res:distortion} disconnects the upper vascular circle from the lower basilar artery, despite showing the smallest mean distortion of 0.055\,mm. 
The morphology is generally better preserved with a large (50\,mm) neighborhood distance, resulting in a nearly planar mesh. 
However, vessel end points are not fully intersected leading to morphological disruptions despite the moderate max.~distortion of 3.34\,mm.
The multi-scale approach (0.5-40\,mm) combines the mentioned advantages, adapting to complex local patterns (with the side effect of having higher max.~distortions), while keeping a low median. 
The distant point pairs consequently prevent the formation of folds.

\subsection{Image-based Auxiliary}

\begin{figure}[tb]
    \centering
    \includegraphics[width=0.81\linewidth]{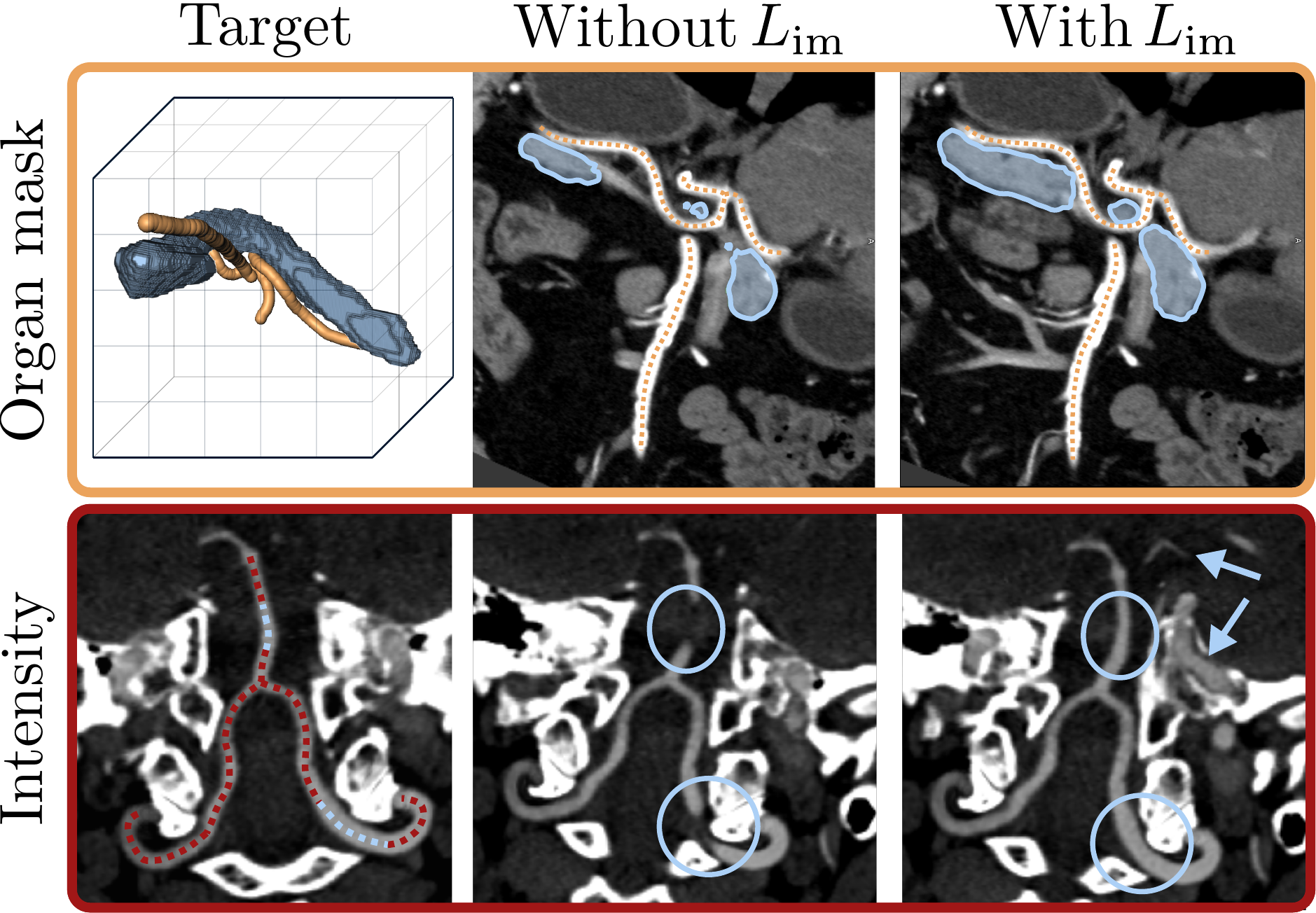}
    \caption{Use cases of the \textbf{image-based loss}. In the top row, we apply $L_i$  to the auxiliary task of displaying the pancreatic organ mask (blue) when unfolding pancreatic vessels. In the bottom row, we unfold the cerebral vasculature, removing (basilar and vertebral) artery segments (blue) from the target annotation (red), which are recovered with the image-based loss (right).}
    \label{fig:res:image}
\end{figure}

In Fig.~\ref{fig:res:image}, we demonstrate two examples involving an image-based loss. 
First, we investigate the retrieval of incomplete or partially incorrect image annotations for missing cerebral vessels. 
For this purpose, we exemplarily remove segments of the \textit{vertebral and basilar artery} from the annotation.
The corresponding loss is modeled as a sink-like function centered around the mean intensity of the annotated vessels, see Sup.~Eq.~\eqref{eq:sup:vessel_sink}.
We recover the missing vessel in the unfolded image (0.44\,mm vs.~2.09\,mm median distance) without image quality loss in the remaining vasculature, while even displaying surrounding vessels (blue arrows). 
Second, we perform unfolding of an auxiliary volumetric target, in this case a pancreas mask.
Here, we aim to unfold as much of the organ mask (instead of its centerline annotation) as possible, along with the primarily targeted abdominal vessels, using a binary mask-based loss given a segmentation mask (see Sup.~Eq.~\eqref{eq:sup:binary_loss}). 
We choose $w_{\text{im}}$\,=\,$10^{-3}$ for both tasks.
Pancreas coverage in the unfolded image doubles from 6.32 to 12.8\,$\text{cm}^2$ without displaying vessel disruptions. 

\subsection{Importance Map \& Sampling}

\begin{figure}[tb]
    \centering
    \includegraphics[width=0.94\linewidth]{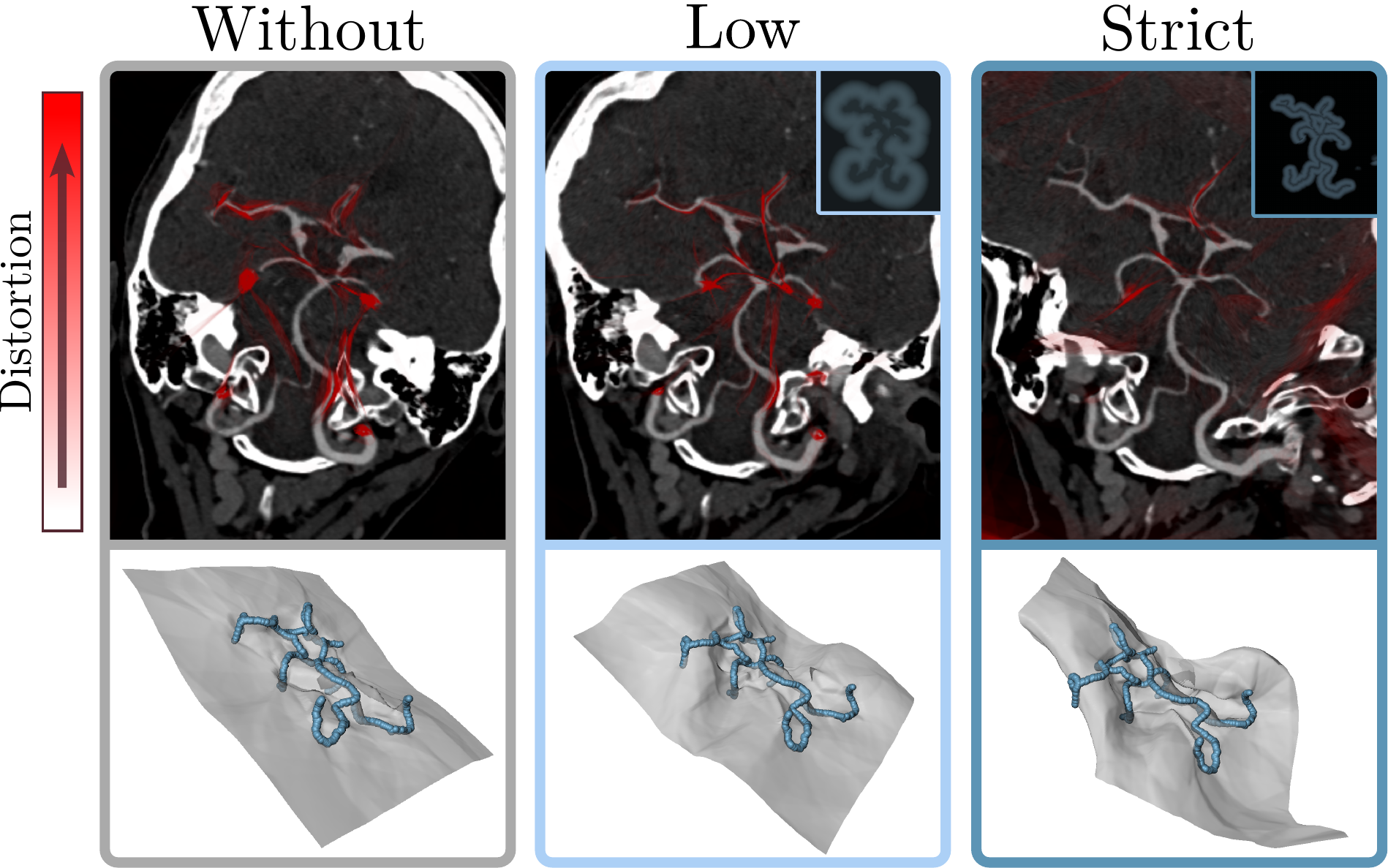}
    \caption{Influence of \textbf{importance map} when fitting cerebral vasculature. Top right corners: no map, weak weighting ($\alpha$=30, $\beta$=0.1) similar to ADR \cite{adr}, strict  weighting ($\alpha$=10, $\beta$=0.1). {\color{red} Pixelwise distortion} is indicated in the unfolded image (top), accompanied by the unfolding parameterization (bottom).} 
    \label{fig:res:importancemap}
\end{figure}

Regarding the influence of importance weights for $L_{\text{d}}$ (see Fig.~\ref{fig:res:importancemap} and Sup.~Tab.~\ref{tab:sup:dist_ablations}), we compare a weak importance map (smooth fade-out of importance values from target locations) and a strict importance map (high weights only close to target). 
Similar to the neighborhood size (cf. Fig.~\ref{fig:res:distortion}), a stricter importance map leads to more flexibility in the image periphery and can hence better adapt to complex structures. 
This is visible in the areal distortion overlay on the right side of the strict importance map in Fig.~\ref{fig:res:importancemap}, where distortion peaks are reduced compared to weak or no weighting (5.35 to 6.06 and 6.45\,mm).
Even with strict weighting, the (mostly unregularized) periphery remains smooth. %

\begin{figure}[tb]
    \centering
    \includegraphics[width=\linewidth]{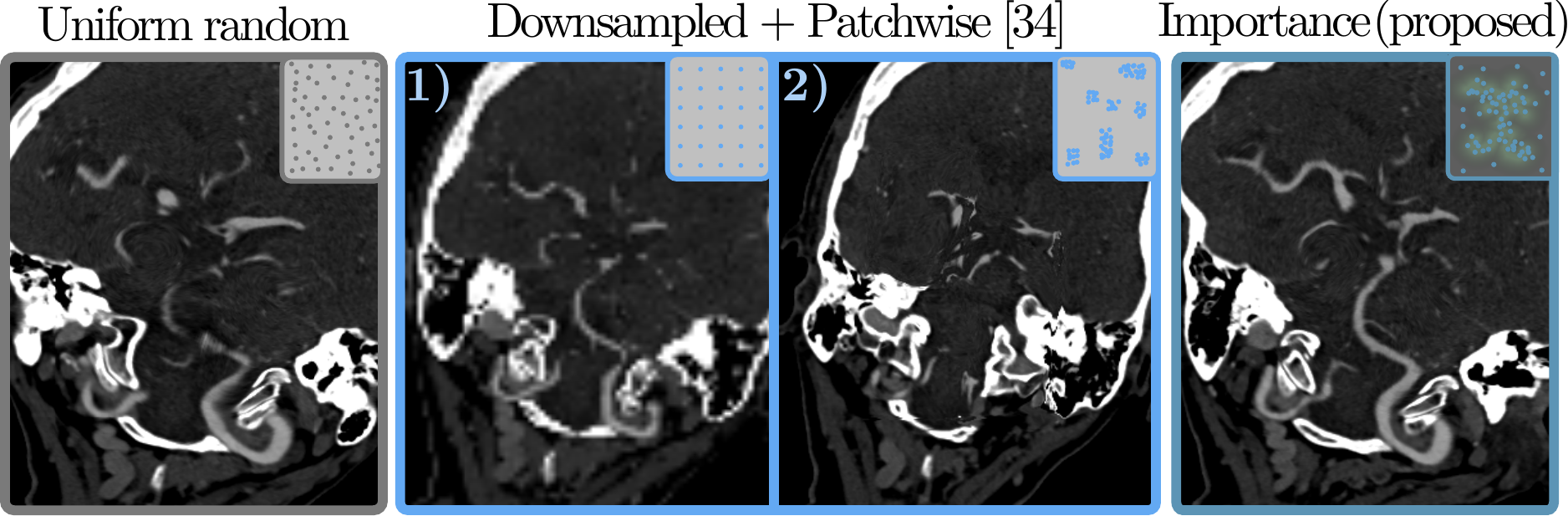}
    \caption{Influence of \textbf{sampling schemes} (in top right corners) in an early fitting stage (500 epochs) when using only 500 samples. Comparing unfinished unfoldings with uniform random (left), hybrid (center) \cite{NIR} and our importance sampling (right).}
    \label{fig:res:importancesampling}
\end{figure}

The influence on convergence of different sampling schemes (uniform-random, downsampled pretraining followed by patchwise fitting \cite{NIR} and our importance sampling) is investigated using 500 samples for the first 500 epochs and is shown in Fig.~\ref{fig:res:importancesampling} and Sup.~Tab.~\ref{tab:sup:sampling_ablation}. 
In early fitting stages, our oversampling of the target regions leads to more consistent centerline coverage ($\approx$ 0.5\,mm improvement on median target distance) with lower median distortion.
In later stages, the effects become marginal compared to uniform-random. %
While downsampled pre-fitting results in a reasonable low-frequency approximation of the target vessels (e.g.~0.87\,mm median distance in Fig.~\ref{fig:res:importancesampling}), the additional patch-wise sampling~\cite{NIR} does not converge to a meaningful solution for this task.
We do not compare against sampling schemes that sample densely in regions with high error \cite{kheradsoftmining2023} as we intend to control the location of distortion.

\section{Discussion}
\subsection{Pipeline}
Our approach shows good morphological coherence, yields generally low maximum distortion values for the complex cerebral vessels and hand bones, and is smooth at distant target points (rib tips, vessel end points) due to its global deformation capabilities. 
For the less complex rib structures with a near-planar configuration, the CeVasMap approach also performs sufficiently well. %
Defining the initial plane size (our sampling space) larger than necessary (twice the size of the target extent) leads to 
smoother transformations as deformations 
can be cushioned by peripheral image parts.
In terms of sampling strategies, the patch-wise approach designed for a fixed grid did not converge, while our approach proves effective in the early stages of the fitting, benefiting potential early-stopping.
Furthermore, our explicit distortion formulation, which evaluates the final point distribution, performs better compared to Jacobian regularizers that operate on the deformation between initial and final mesh. 
Weighting distortion with an importance map additionally allows for flexible handling of complex transformations.

\subsection{Generalization}

Our approach has few requirements compared to prior work since it does not expect any ordering in the target point set compared to, e.g., CPR and is independent of the imaging modality. 
Even though most relevant anatomical targets are locally structured, this enables multi-object scenarios and mitigates potential sources of error.
While unsuitable parameterization of the distortion regularizer may lead to folding which misrepresents the morphology, our multi-scale approach (same parameters for all applications) largely avoids this issue.
The normalization constant $c$, which is crucial for the frequency-embedding to achieve convergence of simpler and smoother manifolds, was set to 750\,mm for the rib cage and 100\,mm for the rest, exhibiting a roughly linear dependence on the target extent.

\subsection{Adaptations \& Flexibility}

We would like to highlight simple ways to adapt, reuse or combine our novel pipeline steps for other use cases. 
Possible advantages of using a pre-deformed (CeVasMap \cite{cevasmap}, ADR \cite{adr}, Sun et al.~\cite{NIR}) or non-linear initial sampling might be practical when specific mesh shapes are preferred or when facing several parallel targets. 
While such a pre-deformation is readily compatible with our approach, we stayed with a single model for the sake of simplicity.
For the image-based loss, more elaborate criteria than pixelwise metrics might be worth investigating since intensities may be shared across different objects, e.g.~vessels and bones, which was circumvented using prior bone removal in our experiments.
Evaluating the overall image appearance can be integrated directly (e.g.~using perceptual losses) if a grid sampling is used.
On a related note, the importance map may also be used to guide the image-based loss to areas distant or close to the target if necessary. 
Lastly, some prior unfolding techniques rely on template-based matching while we prefer to retain the inherent curvature of the anatomy. 
However, distance loss terms between a target point (e.g.~a bifurcation) and a predefined location in the resulting image would also allow template-like constraints.

\subsection{Limitations}
As for all unfolding techniques, the quality of the unfolding relies on the prior (manual or automatic) annotation of target structures of interest from which errors could propagate and wrongly suggest stenosis or missing vessels.
Furthermore, our approach in its current implementation is slower than the ARAP implementation at hand, whose performance mainly depends on the number of points. 
While we trained for 5000 epochs for comparability, suitable states are, however, often already reached earlier when following the early stopping suggestion in Sup.~Sec.~\ref{sec:weighting}.
Most unfoldings are applied for reporting or detailed planning where such durations are not a limitation. 
While the image-based loss can recover structures close to the implicit plane, finding distant structures is unlikely since its capture range for finding other local optima is limited due to its sampling in a 3D subspace.
However, the flexible problem formulation would allow for a more explicit integration (e.g.~using distance losses) if necessary.
This work reports the distortions instead of common Jacobian metrics since these mostly measure deformation from the initial (sub-optimal) mesh.

\section{Conclusion}

Unfolded 2D image representations of 3D anatomical structures improve clinical image visualization, where planar views are unable to capture sparse objects that wind and pass through multiple slices.  %
We present a novel neural-field-based image unfolding pipeline to extract 2-manifolds intersecting these sparse structures. 
For this, we model the non-linear deformation field for an initial PCA-projected mesh as a resolution-independent neural field. %
With this approach, we can effortlessly incorporate flexible geometric regularization terms using point pairs of varying distances and advantageous image-based losses in the optimization.
These can solve auxiliary tasks such as retrieval of objects with missing annotations or inclusion of secondary structures. 
Finally, we show that the fitting process and output quality can be improved by employing importance maps calculated from a distance transform of the targets to guide the point sampling as well as the distortion loss. 
Our modality-independent approach is effective across different medical applications and outperforms baselines especially w.r.t.~maximum distortions for complex structures.
Due to the flexible problem formulation purely based on loss terms and the simple network architecture, our approach may be easily adopted by others in the field or transferred to new applications.

\vspace{1cm}
{
    \small
    \bibliographystyle{ieeenat_fullname}
    \bibliography{main}
}

\clearpage
\setcounter{page}{1}
\setcounter{section}{0}
\maketitlesupplementary

\section{Stopping Criterion} %
\label{sec:weighting}
To assess early convergence, instead of monitoring the decrease of the mutually influencing losses, we propose to stop fitting when the change in the output positions of $G$ uniformly sampled inputs from the current epoch $e_2$ is smaller than $\epsilon$ compared to an earlier epoch $e_1$, i.e.~if 
\begin{equation}
    \sum^G_g ||f^{e_2}(\mathbf{\hat{x}}_g) - f^{e_1}(\mathbf{\hat{x}}_g)|| < \epsilon\,.
\end{equation}

\section{Image Loss formulations}

Loss formulation for the pancreatic task with $V$ being the binary pancreas organ mask:
\begin{equation}
    L_{\text{im}} = \frac{1}{S}\sum^S_s 1 - V(\mathbf{\hat{x}}_s)
    \label{eq:sup:binary_loss}
\end{equation}

\noindent Weighting function for vessel intensity image-loss with $v_{\text{mean}}$ as mean vessel intensity from the target points, $k=0.06$ and $l=200$:
\begin{align}
\begin{split}
    h(\mathbf{\hat{x}}_i) = &~(1 + \exp{(k \cdot (v_{\text{mean}} + l - I(\mathbf{\hat{x}}_i}))))^{-1} \\
    &+ (1 + \exp{(k \cdot (-v_{\text{mean}} + l + I(\mathbf{\hat{x}}_i}))))^{-1}
\end{split}
    \label{eq:sup:vessel_sink}
\end{align}

\newpage

\section{Image Distortion Example}
   
\begin{figure}[!h]
    \centering
    \includegraphics[width=\linewidth]{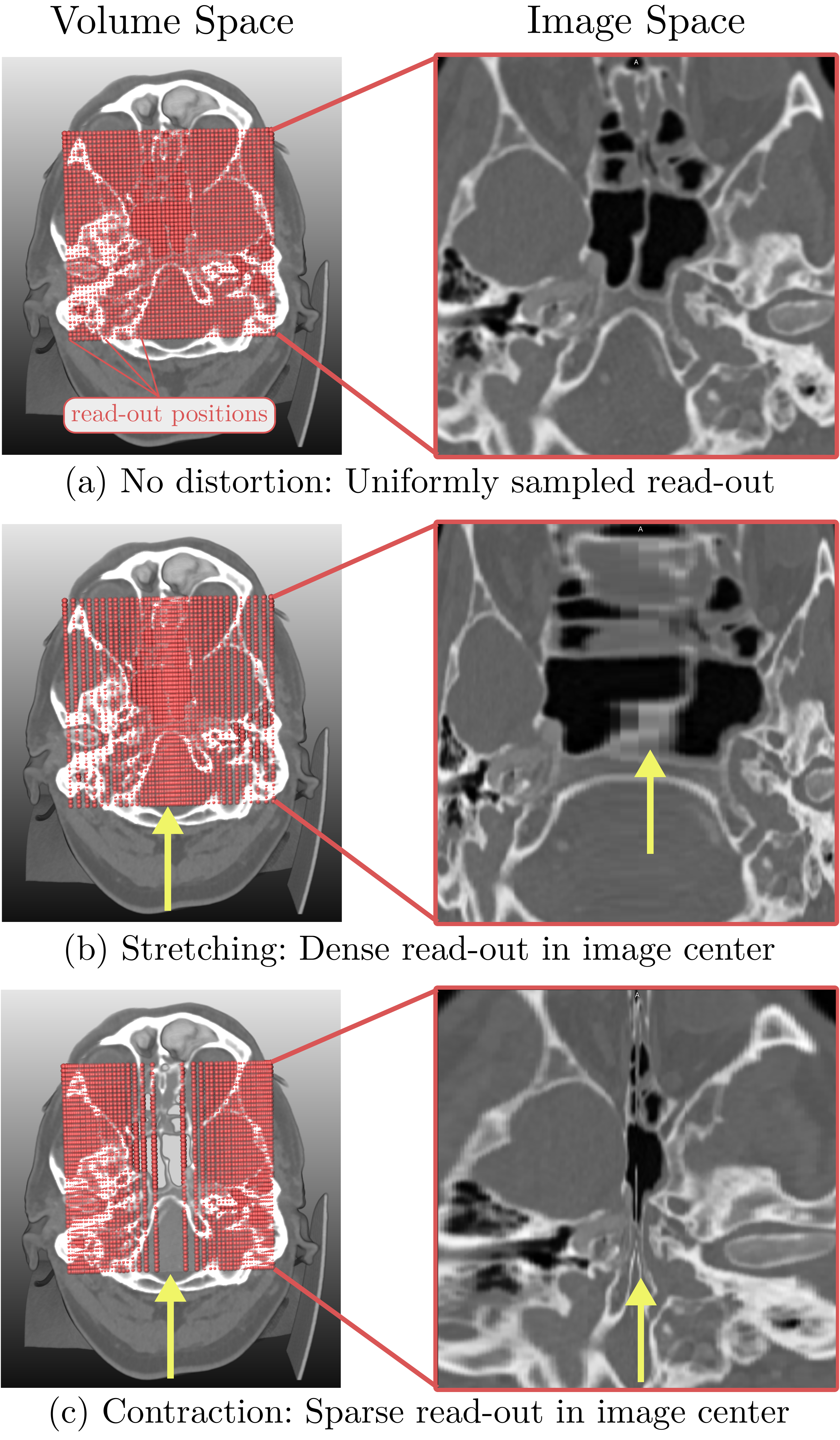}
    \caption{\textbf{Distortion examples} when the distance between neighboring read-out points does not match the image spacing. Left: volumetric readout space with a rendered and cropped head CT in axial direction. Read-out points are displayed in red showing every 8\textsuperscript{th} image row and column. The resulting image is shown on the right. Resulting artifacts in the center of the image are highlighted with yellow arrows.}
    \label{fig:sup:distortionExample}
\end{figure}
\newpage

\onecolumn
\section{Data}

\begin{table}[!htb]
    \centering
    \caption{\textbf{Data specifications} for unfolded structures. The patient data was either publicly available or collected in retrospective
studies which received Institutional Review Board approval. The respective need for informed consent was either given or waived. Rib cage data originated from the public RibSeg data set \cite{ribsegv1}.}
\begin{tabular}{llcc}
    \toprule
        Anatomy (\#) & Included structures & \makecell{Targeted points \\
        (Mean + Std Dev)}& Labeling  \\
        \midrule
        \makecell[c]{Cerebral \\vessels\\ (10)} & \makecell[l]{Vessel centerline points for all important \\ artery surrounding the \textbf{Circle of Willis}: \\Vertebral l./r., basilar, medial (M1+M2),\\ anterior (A1+A2), anterior comm., \\posterior, posterior comm. and ICA C7}& 693 $\pm$ 124 & \makecell[l]{Automatic labeling using \\Rist et al.~\cite{Rist2022BifurcationMF}, \\plus manual correction and\\ addition of M2 + post comm.}\\
        \greyrule
        \makecell[c]{Pancreas \\+ vessels \\(5) }&\makecell[l]{Centerlines of pancreas and \\ the following arteries: \\ Celiac trunk, superior mesenteric, \\ splenic \& common hepatic}& 3214 $\pm$ 245 & \makecell[l]{Skeletonized pancreas segmen-\\ tation (from CNN) and traced \\vessel centerlines following \\bifurcation landmark detection \\(Rist et al.~\cite{pancreatic_vessel})}\\
        \greyrule
        \makecell[c]{Hand bones\\ (5)} & \makecell[l]{Bone centerlines from wrist to \\finger tip for all 5 fingers} & 1384 $\pm$ 73& \makecell[l]{Skeletonized bone annotations \\(from CNN segmentation)} \\
        \greyrule
        \makecell[c]{Rib cage\\ (5)} & Rib centerlines of right rib cage& 3438 $\pm$ 467& \makecell[l]{Provided RibSeg annotations \\were skeletonized}\\
    \bottomrule
    \end{tabular}
    \label{tab:sup:dataset}
\end{table}

\section{Additional Quantitative Results}

\begin{table}[!htb]
\centering
\caption{Comprehensive distortion and distance-to-target overview, both reported in mm. All unfolded images had an isotropic pixel spacing of 0.5\,mm. The reported values are respectively evaluated for 5 rib cages, 5 hand bones, 5 pancreas+vessels and 10 cerebral vessels.}
\begin{tabular}{l  c c c  c c c}
\toprule
 & \multicolumn{3}{c}{\textbf{Hand bones}} & \multicolumn{3}{c}{\textbf{Rib cage}} \\
\cmidrule(lr){2-4} \cmidrule(lr){5-7}
\textbf{Metric} & Ours & ARAP & CeVasMap & Ours & ARAP & CeVasMap \\
\midrule
Distortion Max & 4.736 & 53.188 & 24.574 & 0.269 & 52.715 & 15.160 \\
Distortion Mean & 0.054 & 0.099 &  0.121 & 0.033 & 0.106  & 0.080\\
Distortion Median & 0.028 & 0.021 & 0.047 & 0.027 &   0.029 & 0.042\\
Distortion Std Dev & 0.152 &  0.527 & 0.259 & 0.029 & 0.421 & 0.144\\
\greyrule
Distance Mean  & 0.328 & 0.042 & 0.074 & 0.651 & 0.035 & 0.024 \\
Distance Median  & 0.245 & 0.003 & 0.003&  0.479 & 0.000 & 0.000\\
Distance Std Dev & 0.379 & 0.129 & 0.270& 0.597 & 0.150 & 0.131\\
\midrule
 & \multicolumn{3}{c}{\textbf{Pancreas + vessels}} & \multicolumn{3}{c}{\textbf{Cerebral vessels}} \\
\cmidrule(lr){2-4} \cmidrule(lr){5-7}
 & Ours & ARAP & CeVasMap & Ours & ARAP & CeVasMap \\
 \midrule
Distortion Max & 0.829 & 19.714 & 11.642 & 6.057 &22.568  & 22.007\\
Distortion Mean & 0.047 & 0.066 & 0.096 & 0.082 & 0.082 & 0.117\\
Distortion Median & 0.026 & 0.022 & 0.051 & 0.034 & 0.018 & 0.048\\
Distortion Std Dev & 0.065  & 0.202 & 0.210 & 0.169 & 0.297 & 0.297\\
\greyrule
Distance Mean & 0.253 & 0.138 & 0.136 & 0.352 & 0.040 & 0.071\\
Distance Median &  0.234 & 0.101 & 0.101 & 0.306 & 0.000 & 0.000\\
Distance Std Dev & 0.154 & 0.131 & 0.117 & 0.206 & 0.149 & 0.234\\
\bottomrule
\end{tabular}
\label{tab:sup:baseline}
\end{table}

\begin{table}[!htb]
\centering
\caption{Distortion values of 10 cerebral vessel trees, evaluated on images with 0.5\,mm isotropic pixel spacing. Distortion was computed per neighboring pixel pair and is given in mm. Table compares multiple \textbf{distortion regularizers} as well as the different \textbf{importance map} settings against a baseline using a weak importance map and a multi-scale distortion regularizer.}
\begin{tabular}{lccccccc}
\toprule
 & \textbf{Baseline} &\multicolumn{4}{c}{\textbf{Distortion Reg}} & \multicolumn{2}{c}{\textbf{Importance Map}} \\
\cmidrule(lr){2-2} \cmidrule(lr){3-6} \cmidrule(lr){7-8}
 \textbf{Metric}& MS $\vert$ Weak & $J_1$ \cite{wolterink2022implicit} & $J_2$ \cite{NIR} & Ours 10\,mm  & Ours 50\,mm & No & Strict \\
\midrule
Distortion Max &  6.057 &0.957& 1.384 & 3.962 & 3.338 & 6.445 &5.354 \\
Distortion Mean &  0.082 &0.125& 0.146 & 0.055 & 0.119 & 0.070 & 0.071\\
Distortion Median &  0.034& 0.090& 0.101 & 0.032 &0.047 & 0.027  & 0.034\\
Distortion Std Dev &   0.169 &0.114& 0.149 & 0.090 & 0.209 & 0.155 & 0.131\\
\bottomrule
\end{tabular}
\label{tab:sup:dist_ablations}
\end{table}

\begin{table}[!htb]
\centering
\caption{Distortion and target distance values of 10 cerebral vessel trees, evaluated on images with 0.5\,mm isotropic pixel spacing. Distortion was computed per neighboring pixel pair and is given in mm. Distance values are computed as closest distance to mesh for each target point in mm. 
Table compares multiple \textbf{sampling schemes}.}
\begin{tabular}{lcccc}
\toprule
\textbf{Metric} & \textbf{Uniform} & \textbf{Hybrid}~\cite{NIR} & \textbf{Importance} \\
\midrule
Distortion Mean & 0.129 & 0.106  & 0.093 \\
Distortion Median & 0.102  & 0.039 & 0.065\\
Distortion Std & 0.110 & 0.190 & 0.099\\
\greyrule
Distance Mean & 4.162 & 4.339 & 3.886\\
Distance Median &  2.981 & 2.856 & 2.515 \\
Distance Std Dev & 3.935 & 4.298 & 4.059 \\
\bottomrule
\end{tabular}
\label{tab:sup:sampling_ablation}
\end{table}

\begin{figure}[!htb]
    \centering
    \includegraphics[width=\linewidth]{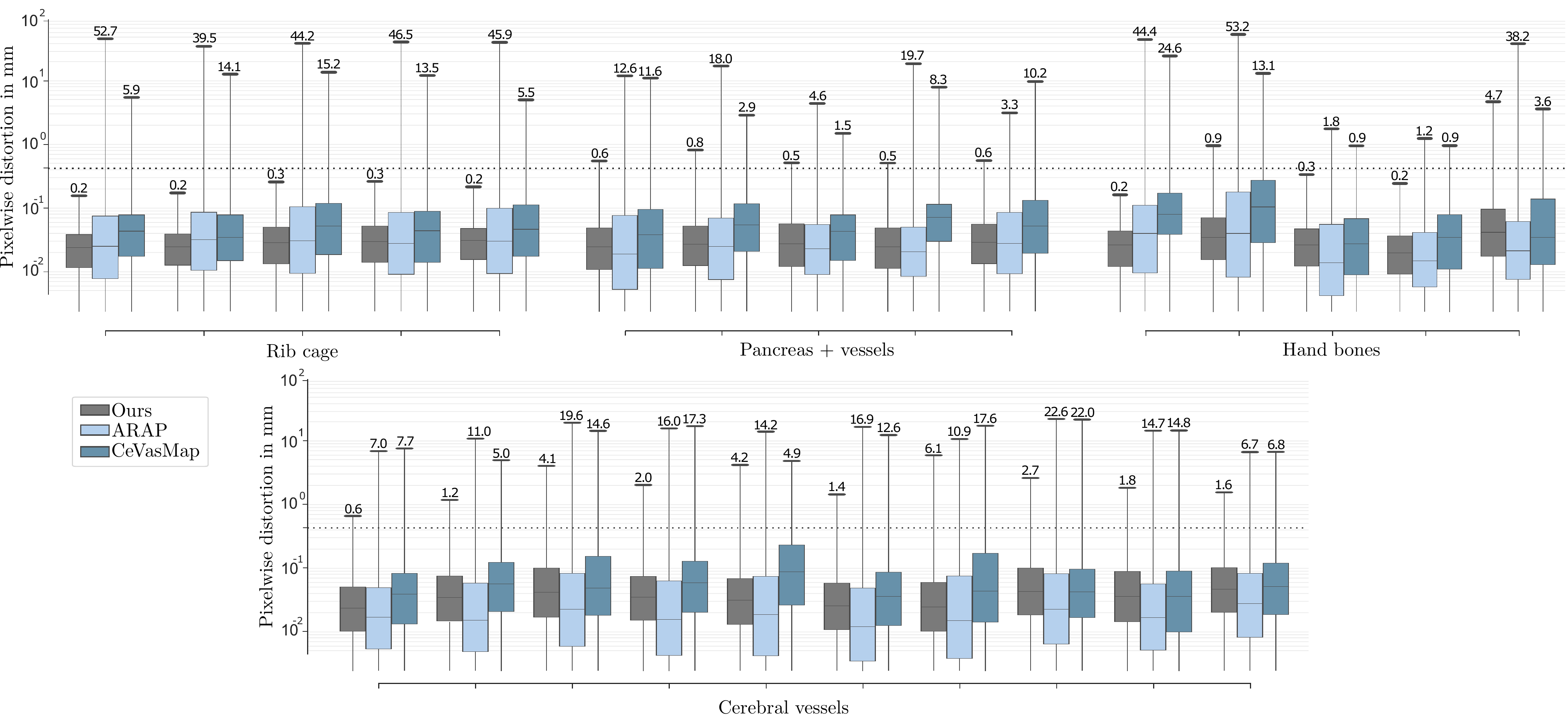}
    \caption{\textbf{Logarithmic distortion distribution} as boxplot (median line, std-dev.~boxes, maximum whiskers) within 1\,cm radius around unfolded target for each case per anatomy and method. Horizontal line (dotted) of pixel spacing $\delta$, approx.~indicates the beginning of severe distortions. This plot shows the per-case distortion from the aggregation in Fig.~\ref{fig:res:dist_plot}.}
    \label{fig:enter-label}
\end{figure}
\twocolumn

\end{document}